\documentclass[runningheads]{llncs}
\usepackage[T1]{fontenc}
\usepackage{times}
\usepackage[hyphens]{url}
\usepackage{algorithm}
\usepackage{algorithmic}
\usepackage{newfloat}
\usepackage{listings}
\usepackage{amsmath}
\usepackage{amssymb}
\usepackage{graphicx}
\usepackage{xspace}
\usepackage{xcolor}
\usepackage[misc]{ifsym}
\newcommand{\our}{\text{Bridge}\xspace}
\usepackage{subfig}
\begin{document}
\title{Bridge Structural Knowledge and Pre-trained Language Models for Knowledge Graph Completion}
\titlerunning{Bridge Knowledge Graph Completion}
%
\author{Qiao Qiao$^*$ \Letter \and
  Yuepei Li$^*$ \and
  Qing Wang \and
  Kang Zhou \and
  Qi Li\orcidID{0000-0002-3136-2157}
}

\authorrunning{Q. Qiao et al.}
%
\institute{Iowa State University, Ames, Iowa, USA
\email{\{qqiao1,liyp0095,qingwang,kangzhou,qli\}@iastate.edu}}
%
\maketitle              
\def\thefootnote{*}\footnotetext{These authors contributed equally to this work.}
%
\begin{abstract}
Knowledge graph completion (KGC) is a task of inferring missing triples based on existing Knowledge Graphs (KGs). Both structural and semantic information are vital for successful KGC. However, existing methods only use either the structural knowledge from the KG embeddings or the semantic information from pre-trained language models (PLMs), leading to suboptimal model performance. Moreover, since PLMs are not trained on KGs, directly using PLMs to encode triples may be inappropriate. To overcome these limitations, we propose a novel framework called \our, which jointly encodes structural and semantic information of KGs. Specifically, we strategically encode entities and relations separately by PLMs to better utilize the semantic knowledge of PLMs and enable structured representation learning via a structural learning principle. Furthermore, to bridge the gap between KGs and PLMs, we employ a self-supervised representation learning method called BYOL to fine-tune PLMs with two different views of a triple. Unlike BYOL, which uses augmentation methods to create two semantically similar views of the same image, potentially altering the semantic information. We strategically separate the triple into two parts to create different views, thus avoiding semantic alteration. Experiments demonstrate that \our outperforms the SOTA models on three benchmark datasets.

\end{abstract}

\section{Introduction}\label{sec:intro}

Knowledge graphs (KGs) are graph-structured databases composed of triples (facts), where each triple $(h, r, t)$ represents a relation $r$ between a head entity $h$ and a tail entity $t$. KGs such as Wikidata \cite{vrandevcic2014wikidata} and WordNet \cite{fellbaum1998wordnet} have a significant impact on various downstream applications such as named entity recognition \cite{li2025re,zhou2022distantly}, relation extraction \cite{wang2023improving}, and entity linking \cite{zhou2024gendecider}. Nevertheless, the effectiveness of KGs has long been hindered by the challenge of the incompleteness problem.

To address this issue, researchers have proposed a task known as Knowledge Graph Completion (KGC), which aims to predict missing relations and provides a valuable supplement to enhance KG's quality. Most existing KGC methods fall into two main categories: structure-based and pre-trained language model (PLMs)-based methods. Structure-based methods represent entities and relations as low-dimensional continuous embeddings, which effectively preserve their intrinsic structure \cite{bordes2013translating,dettmers2018convolutional,ge2023compounding,kim2022bootstrapped}. While effective in KG's structure representation learning, these methods overlook the semantic knowledge associated with entities and relations. Recently, PLM-based models have been proposed to leverage the semantic understanding captured by PLMs, adapting KGC tasks to suit the representation formats of PLMs \cite{kim2020multi,qiao2023improving,wang2021structure,wang2022simkgc,yao2019kg}. 

While these models offer promising potential to enhance KGC performance, there is space to improve: (1) Existing structure-based methods do not explore knowledge provided by PLMs. (2) Existing PLM-based methods aim to convert KGC tasks to fit language model format and learn the relation representation from a semantic perspective using PLMs, overlooking the context of the relation in KGs. Consequently, they lack optimal alignment with structural knowledge. For example, given a triple \emph{(trade name, member of domain usage, metharbital)}\footnote{This is a triple from WordNet, and metharbital is an anticonvulsant drug used in the treatment of epilepsy.}, the semantic of the relation \emph{member of domain usage} is ambiguous since ``it is not a standard used term in the English\footnote{interpretation from ChatGPT when asking ``what does \emph{member of domain usage} mean?'' }''; hence, PLMs may lack accurate semantic representation. Thus, it becomes imperative to enable the model to leverage the principle of structural learning to grasp structural knowledge and compensate for the limitations of semantic understanding. (3) Existing PLM-based methods utilize PLMs directly, overlooking the disparity between PLMs and triples arising from the lack of triple training during PLMs pre-training. This oversight limits the expressive power of PLMs and their adaption to the KG's domain. 

To address the limitations of existing methods, we propose a two-in-one framework named \our. To overcome the challenge of lacking structural knowledge in PLMs, we propose a structured triple knowledge learning phase. Specifically, we follow the widely applied principle in traditional structured representation learning for KGs \cite{balazevic2019tucker,bordes2013translating,qiao2023relation,shang2024mixed,sun2018rotate}, which posits that the relation is a translation from the head entity to the tail entity. We strategically extract the embedding of $h,r$ and $t$ separately from PLMs and employ various structure-based scoring functions to assess the plausibility of a triple. This approach allows us to reconstruct KG's structure in the semantic embedding via the structured learning principle. This principle has been widely applied in traditional structured representation learning for KGs, but there is no previous study that investigates this principle using PLM-based representation.

However, due to the different principles between traditional structured representation learning and PLMs, there is a gap between them since PLMs are not trained on KGs. To bridge the gap between PLMs and KGs, we fine-tune PLMs to integrate structured knowledge from KGs into PLMs. 
By taking this step, we unify the space of structural and semantic knowledge, making integration of KGs and PLMs more reasonable. 
In summary, our main contributions are:
    \begin{enumerate}
        \item We propose a general framework, \our, that jointly encodes structural and semantic information of KGs and can incorporate various scoring functions. 
        \item We utilize BYOL innovatively for fine-tuning PLM to bridge the gap between structural knowledge and PLMs. 
        \item We conduct empirical studies with two widely used structural-based scoring functions on three benchmark datasets. Experiment results show that \our consistently and significantly outperforms other baseline methods.
    \end{enumerate}

\section{Related Work}\label{sec:related}

\subsection{Structure-based KGC}
Structure-based KGC aims to embed entities and relations into a low-dimensional continuous vector space while preserving their intrinsic structure through the design of different scoring functions. Various knowledge representation learning methods can be divided into the following categories: (1) Translation-based models, which assess the plausibility of a fact by calculating the Euclidean distance between entities and relations \cite{bordes2013translating,ge2023compounding,ji2015knowledge,sun2018rotate}; (2) Semantic matching-based models, which determine the plausibility of a fact by calculating the semantic similarity between entities and relations \cite{balazevic2019tucker,liang2023knowledge,nickel2011three,yang2015embedding}; and (3) Neural network-based models, which employ deep neural networks to fuse the graph network structure and content information of entities and relations \cite{guan2018shared,kim2022bootstrapped,qiao2023relation,shang2024mixed,shang2019end,vashishth2019composition}. 
All these structure-based models are limited to using graph structural information from KGs, and they do not leverage the rich contextual semantic information of PLMs to enrich the representation of entities and relations.

\subsection{PLM-based KGC}
PLM-based KGC refers to a method for predicting missing relations in KGs using the implicit knowledge of PLMs. KG-BERT \cite{yao2019kg} is the first work to utilize PLMs for KGC. It treats triples in KGs as textual sequences and leverages BERT \cite{kenton2019bert} to model these triples. MTL-KGC \cite{kim2020multi} utilizes a multi-task learning strategy to learn more relational properties. This strategy addresses the challenge faced by KG-BERT, where distinguishing lexically similar entities is difficult. To improve the inference efficiency of KG-BERT, StAR \cite{wang2021structure} partitions each triple into two asymmetric parts and subsequently constructs a bi-encoder to minimize the inference cost. SimKGC \cite{wang2022simkgc} proposes to utilize contrastive learning to improve the discriminative capability of the learned representation. Adopting the architecture of SimKGC, GHN \cite{qiao2023improving} develops an innovative self-information-enhanced contrastive learning approach to generate high-quality negative samples. MPIKGC \cite{xu2024multi} utilizes large language models (LLMs) to enrich the descriptions of entities/relations.
In contrast to previous encode-only models, \cite{chen2022knowledge,saxena2022sequence} explore the generation-based models that directly generate a target entity.
However, all these methods simply involve fine-tuning PLMs directly, disregarding both the absence of structured knowledge in PLMs and the gap between PLMs and KGs.

\section{Preliminary} \label{sec:preliminary}

\subsection{Bootstrap Your Own Latent (BYOL)} \label{section:BYOL}

Bootstrap Your Own Latent (BYOL) is an approach to self-supervised image representation learning without using negative samples. It employs two networks, referred to as the \emph{online} and \emph{target} networks, working collaboratively to learn from one another.  
The \emph{online} network is defined by a set of weights $\theta$, while the \emph{target} network shares the same architecture as the \emph{online} network but utilizes a different set of weights $\xi$. 

Given the image $x$, BYOL generates two augmented views $(v,v')$ from the image $x$ using different augmentations. These two views $(v,v')$ are separately processed by the \emph{online} and the \emph{target} encoders. The \emph{online} network produces a representation $\mathbf{y_\theta} = f_\theta(v)$ and a projection $\mathbf{z_\theta}=g_\theta(\mathbf{y_\theta})$, while the \emph{target} network outputs a representation $\mathbf{y'_\xi} = f_\xi(v')$ and a projection $\mathbf{z'_\xi}=g_\xi(\mathbf{y'_\xi})$. Next, only the \emph{online} network applies a prediction $q_\theta(\mathbf{z_\theta})$, creating an asymmetric between the \emph{online} and the \emph{target} encoders. Finally, the loss function is defined as the mean squared error between the normalized predictions and target projections :
\begin{equation}\label{eq:byol1}
    \mathcal{L}_{\theta,\xi}\triangleq\| \bar{q_\theta}(\mathbf{z_\theta})-\mathbf{\bar{z}'_{\xi}}\|^2_2 = 2 - 2 \cdot \frac{\langle q_\theta(\mathbf{z_\theta}),\mathbf{z'_\xi\rangle}}{\|{q_\theta(\mathbf{z_\theta})\|}_2 \cdot \mathbf{\|z'_\xi\|}_2},
\end{equation}
where $\bar{q_\theta}(\mathbf{z_\theta})$ and $\mathbf{\bar{z}'_{\xi}}$ are the $l2$-normalized term of $q_\theta(\mathbf{z_\theta})$ and $\mathbf{z'_{\xi}}$.

To symmetrize the loss $ \mathcal{L}_{\theta,\xi}$, BYOL swaps the two augmented views of each network, feeding $v'$ to the \emph{online} network and $v$ to the \emph{target} network to compute $\widetilde{L}_{\theta,\xi}$. During each training step, BYOL performs a stochastic optimization step to minimize $\mathcal{L}^{BYOL}_{\theta,\xi} = \mathcal{L}_{\theta,\xi} + \widetilde{L}_{\theta,\xi}$ with respect to $\theta$ only. $\xi$ are updated after each training step using an exponential moving average of the online parameters $\theta$ as follows:
\begin{equation}\label{eq:byol2}
\begin{gathered}
    \xi \leftarrow \tau \xi + (1 - \tau) \theta,
    \end{gathered}
\end{equation}
where $\tau$ is a target decay rate.

\vspace{-0.1in}
\subsection{Problem Definition}


\paragraph{Knowledge Graph Completion}
The knowledge graph completion (KGC) task is to either predict the tail/head entity $t/h$ given the head/tail entity $h/t$ and the relation $r$: $(h,r,?)$ and $(?,r,t)$, or predict relation $r$ between two entities: $(h,?,t)$. In this work, we focus on head and tail entity prediction.

\section{Methodology}\label{sec:methodology}

\begin{figure}[t]
\centering
\subfloat[\centering\label{fig:1(a)} ]
{{\includegraphics[width=7cm]{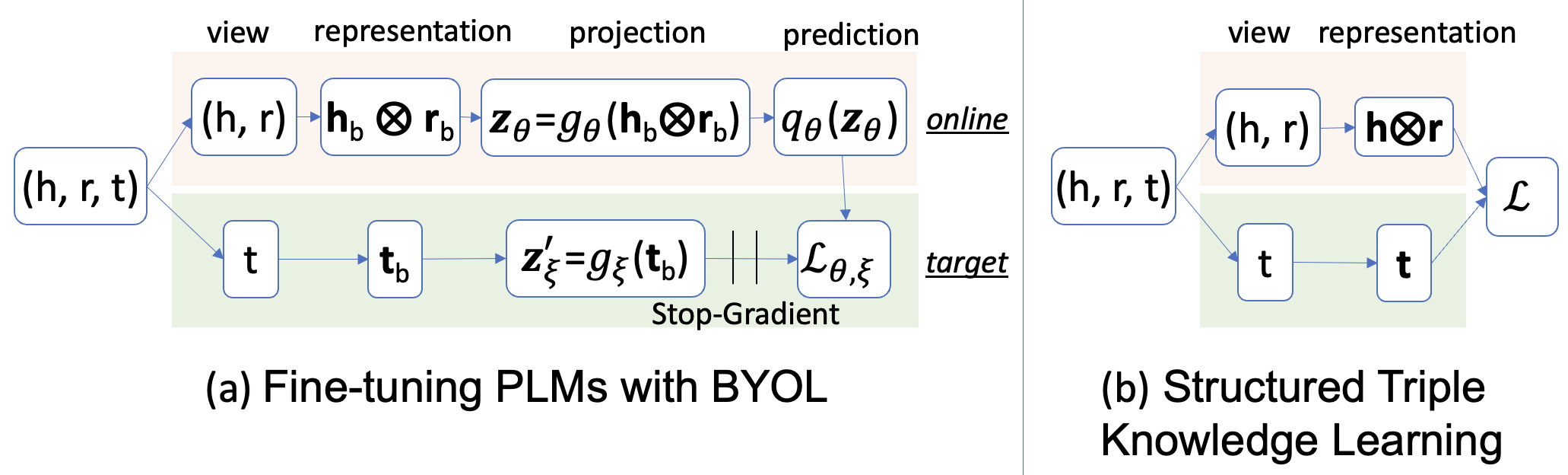}}}%
\qquad
\subfloat[\centering\label{fig:1(b)} ]{{\includegraphics[width=4cm]{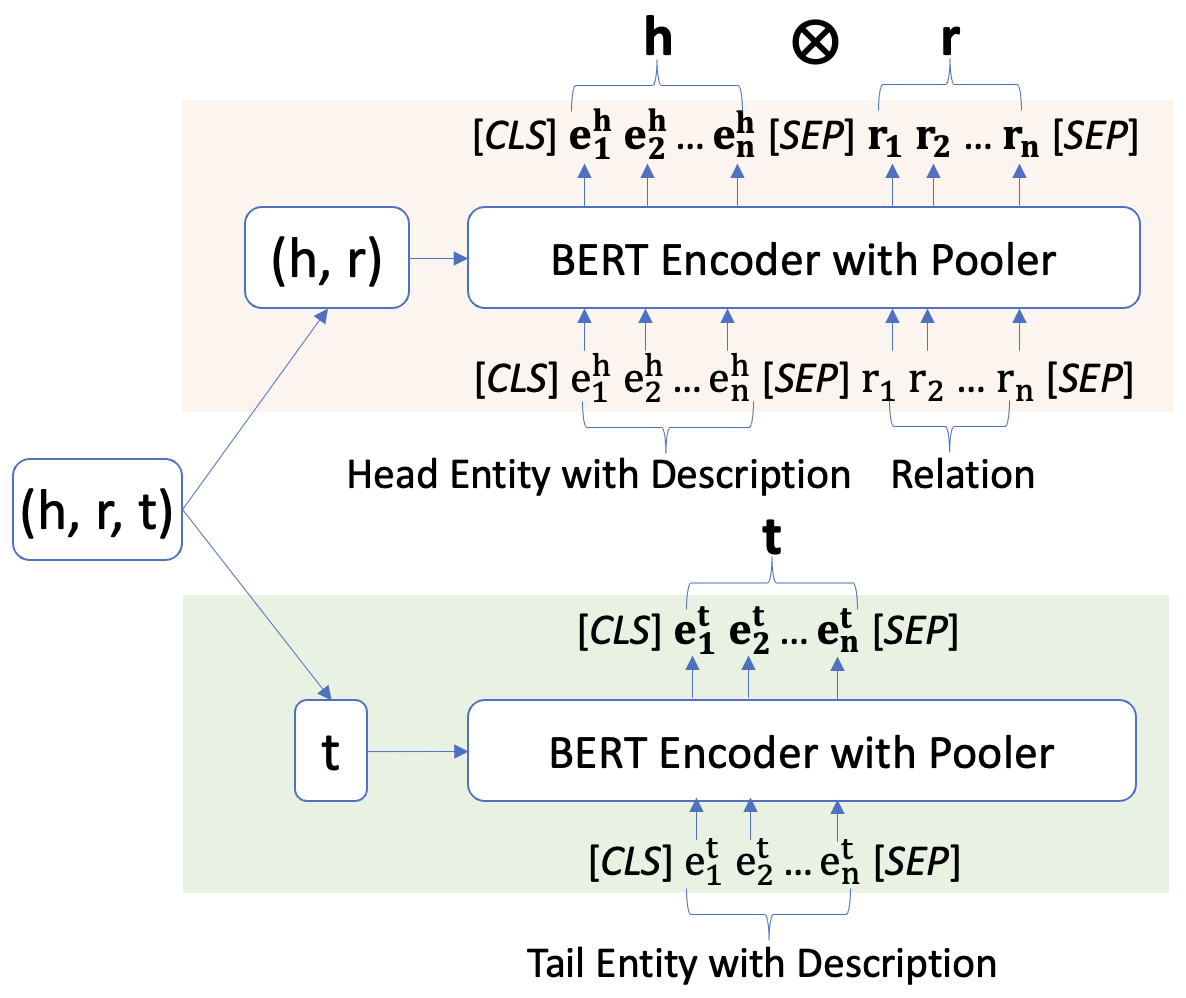} }}%
\caption{(a) The framework of \our. $\otimes$ represents different interaction strategies between entities and relations, determined by various scoring functions. (b) Structure-Aware PLM Encoder.}
\vspace{-0.2in}
\label{fig:framework}
\end{figure}

In this section, we present \our in detail. We first introduce a structure-aware PLM encoder, which aims to learn structure knowledge by PLMs. Then we introduce two essential modules in \our. The first module utilizes a fine-tuning process with BYOL to seamlessly integrate structural knowledge from KGs into PLMs, thereby bridging the gap between the two. The second module aims to learn structure-enhanced triple knowledge with PLMs, allowing PLMs to acquire domain knowledge of KGs.
As shown in Fig.\ref{fig:1(a)}, \our integrates these two modules by sequentially training two objectives.
We take the tail entity prediction task $(h,r,?)$ as an example to illustrate the procedure, and the procedure for the head entity prediction task $(?,r,t)$ is the same.

\vspace{-0.1in}

\subsection{Structure-Aware PLMs Encoder}\label{section:Structure-Aware PLMs Encoder}
Existing structure-based and PLM-based methods can lead to suboptimal performance, especially when dealing with ambiguous relations. Hence, it is essential to incorporate structural knowledge with semantic knowledge to achieve a structure-enhanced relation representation.
To facilitate structure representation learning, we use two BERT encoders. Given a triple $(h,r,t)$, the first encoder takes the textual description of the head entity $h$ and relation $r$ as input, where the textual description of the head entity $h$ is denoted as  $(e_1^h,e_2^h,\cdots,e_n^h)$, and relation $r$ is denoted as a sequence of tokens $(r_1,r_2,\cdots,r_n)$, the input sequence is: $[CLS]\ e_1^h \ e_2^h \ \cdots \ e_n^h \ [SEP]\ r_1 \ r_2 \ \cdots \ r_n\ [SEP]$. 
The second encoder takes the textual description of the tail entity $t$ as input, where the textual description of the tail entity $t$ is denoted as a sequence of tokens $(e_1^t,e_2^t,\cdots,e_n^t)$, the input sequence format is: $[CLS]\ e_1^t \ e_2^t\ \cdots\ e_n^t \ [SEP]$. The design of these two encoders is illustrated in Fig.\ref{fig:1(b)}. The embedding of $h,r,t$ is computed by taking the mean pooling of the corresponding BERT output:
\begin{equation}\label{eq:pooling}
\begin{gathered}
    \mathbf{h} = MeanPooling(\mathbf{e_1^h},\mathbf{e_2^h},\cdots,\mathbf{e_n^h}),\\
    \mathbf{r} = MeanPooling(\mathbf{r_1},\mathbf{r_2},\cdots,\mathbf{r_n}),\\
    \mathbf{t} = MeanPooling(\mathbf{e_1^t},\mathbf{e_2^t},\cdots,\mathbf{e_n^t}).
\end{gathered} 
\end{equation}



To reconstruct KG's structure in the semantic embedding, we analyze two widely applied scoring functions in the KGC task, including TransE 
and RotatE. The corresponding structure scoring functions $\phi(h,r,t)$ are designed as follows:
\begin{equation}\label{eq:score_transE}
    \phi(h,r,t) = \phi(h \otimes r,t)_{TransE} = \cos(\mathbf{h+r},\mathbf{t}) = \frac{\mathbf{(h+r)} \cdot \mathbf{t}}{\|\mathbf{h+r}\| \|\mathbf{t}\|}.
\end{equation}

\begin{equation}\label{eq:score_rotatE}
    \phi(h,r,t) = \phi(h \otimes r,t)_{RotatE} = \cos(\mathbf{h \circ r},\mathbf{t}) = \frac{\mathbf{(h \circ r)} \cdot \mathbf{t}}{\|\mathbf{h \circ r}\| \|\mathbf{t}\|}.
\end{equation}

where $\circ$ denotes the Hadamard (element-wise) product, and $\otimes$  represents different interaction strategies between entities and relations.
Note that \our is flexible enough to be generalized to other existing structure-based scoring functions.










\vspace{-0.1in}
\subsection{Fine-tuning PLMs with BYOL} \label{section:Fine-tuning}

Previous PLM-based approaches leverage PLMs directly and disregard the gap between structure knowledge and PLMs because PLMs are not trained on triples. Therefore, strategic fine-tuning PLMs is necessary. Considering the existence of one-to-many, many-to-one, and many-to-many relations in KGs, 
we exclusively consider positive samples and adopt BYOL \cite{grill2020bootstrap} as it does not require negative samples. We generate an alternative view of KG by separating a triple into two parts, and leveraging the widely used structural principles to learn KG information. 

BYOL generates two augmented views of the same instance, with one view serving as the input for the \emph{online} network and the other view serving as the input for the \emph{target} network. 
Here, the \emph{online} encoder takes the textual descriptions of the head entity $h$ and relation $r$ as input and produces an \emph{online} representation $\mathbf{h_b \otimes r_b}$. The \emph{target} encoder takes the textual descriptions of the tail entity $t$ as input and produces a \emph{target} representation $\mathbf{t_b}$. The design of the encoder is elaborated in Section \ref{section:Structure-Aware PLMs Encoder}. 

The \emph{online} projection network $g_\theta$ takes the \emph{online} representation $\mathbf{h_b \otimes r_b}$ as input and outputs an \emph{online} projection representation $\mathbf{z_\theta}$:
\begin{equation}   \mathbf{z_\theta}=g_\theta(\mathbf{h_b \otimes r_b}) = \mathbf{W_2[\sigma(W_1[h_b \otimes r_b])]},
\end{equation}
where $\mathbf{W_1}$ and $\mathbf{W_2}$ are trainable parameters, $g_\theta$ is a MLP network with one hidden layer, $\sigma(\cdot)$ is a PReLU function, and $\otimes$  represents different interaction strategies between entities and relations, determined by various scoring functions.

The \emph{target} projection network $g_\xi$ takes the \emph{target} representation $\mathbf{t_b}$ as input and outputs a \emph{target} projection representation $\mathbf{z'_\xi}$:
\begin{equation}
    \mathbf{z'_\xi}=g_\xi(\mathbf{t_b}) = \mathbf{W_4[\sigma(W_3t_b)]},
\end{equation}
where $\mathbf{W_3}$ and $\mathbf{W_4}$ are trainable parameters, $g_\xi$ is a MLP network with one hidden layer, and $\sigma(\cdot)$ is a PReLU function.

The prediction network $q_\theta$ takes the \emph{online} projection representation $\mathbf{z_\theta}$ as input and outputs a representation $q_\theta(\mathbf{z_\theta})$ which is a prediction of the \emph{target} projection representation $\mathbf{z'_\xi}$, the goal is to let the \emph{online} network predict the \emph{target} network's representation of another augmented view of the same triple:
\begin{equation}
    q_\theta(\mathbf{z_\theta}) \approx \mathbf{z'_\xi},
\end{equation}
where $q_\theta$ is a MLP network with one hidden layer.


\subsection{Structured Triple Knowledge Learning}\label{section:structure}





To reconstruct KG's structures in the semantic embedding, after fine-tuning PLMs with BYOL, we employ the fine-tuned \emph{online} encoder and the \emph{target} encoder to facilitate structure learning. The \emph{online} BERT encoder takes the textual description of the head entity $h$ and the relation $r$ as input. The \emph{target} BERT encoder takes the textual description of the tail entity $t$ as input. The structure scoring function $\phi(h,r,t)$  is utilized to train these two encoders further to incorporate structure knowledge into PLMs.

\subsection{Objective and Training Process}

During the Fine-tuning PLMs with BYOL phase, we optimize the PLMs for domain adaption in KGs using the loss $\mathcal{L}_{\theta,\xi}$,  which is computed according to Eq.\eqref{eq:byol1}. The \emph{online} parameters $\theta$ are updated by a stochastic optimization step to make the predictions $q_\theta(\mathbf{z_\theta})$ closer to  
$\mathbf{z'_\xi}$ for each triple, while the target parameters $\phi$ are updated by Eq.\eqref{eq:byol2}. To symmetrize this loss, we also swap the input of the \emph{online} and \emph{target} encoder. 

In the Structured Triple Knowledge Learning phase, we use contrastive loss with additive margin \cite{wang2022simkgc} to simultaneously optimize the structure and PLMs objectives: 
\begin{equation}
    \mathcal{L} = -log\frac{e^{(\phi(h,r,t)-\gamma)/\tau}}{e^{(\phi(h,r,t)-\gamma)/\tau}+\sum_{i=1}^{|\mathcal{N}|} e^{(\phi(h,r,t'_i)-\gamma)/\tau}},
\end{equation}
where $\tau$ denotes the temperature parameter, $t'_i$ denotes the $i$-th negative tail, $\phi(h,r,t)$ is the score function as in Eq.\eqref{eq:score_transE}, Eq.\eqref{eq:score_rotatE}, and the additive margin $\gamma > 0$ encourages the model to increase the score of the correct triple $(h,r,t)$. 

\section{Experimental Study}\label{sec:exper}


\subsection{Datasets and Evaluation Metrics}

We run experiments on three datasets: WN18RR \cite{fellbaum1998wordnet}, FB15k-237 \cite{toutanova2015representing}, and Wikidata5M \cite{wang2021kepler}. 
The statistics are shown in Table \ref{table:dataset}. We employ two evaluation metrics: Hits@K and mean reciprocal rank. Hits@K indicates the proportion of correct entities ranked in the top $k$ positions, while MRR represents the mean reciprocal rank of correct entities. 

\begin{table}
\caption{Statistics of the Datasets. Columns 2-6 represent the number of entities, relations, triples in the training set, validation set, and the test set, respectively.}\label{table:dataset}
\centering
\begin{tabular}{c|c|c|c|c|c} 
 \hline
 Dataset & \#Ent & \#Rel & \#Train & \#Valid  & \#Test  \\
 \hline
 WN18RR & $40,943$  & $11$ & $86,835$ & $3,034$& $3,134$ \\
 FB15k-237 & $14,541$ & $237$  & $272,115$& $17,535$& $20,466$\\
 Wikidata5M-Trans & $4,594,485$  & $822$ & $20,614,279$ & $5,133$& $5,163$ \\
 \hline
\end{tabular}
\end{table}

\begin{table*}[t]  
\centering
  \caption{Main results. \textbf{Bold} represents the best results and \underline{underline} denotes the runner-up results, $\dagger$ cites the results from \cite{wang2022simkgc}, $*$ cites the results from original papers. - indicates that the original papers do not present results related to the corresponding dataset.}
\resizebox{\textwidth}{!}{%
  \begin{tabular}{l||cccc|cccc|cccc}
  \hline
     & \multicolumn{4}{c|}{WN18RR}   & \multicolumn{4}{c|}{FB15k-237}  & \multicolumn{4}{c}{Wikidata5M-Trans} \\
      \cline{2-13}
      Model & {MRR} & {Hits@1} & {Hits@3} & {Hits@10} & {MRR} & {Hits@1} & {Hits@3} & {Hits@10} & {MRR} & {Hits@1} & {Hits@3} & {Hits@10}\\
      \hline
      \multicolumn{13}{c}{Structure-based Methods} \\
      \hline
    TransE$^\dagger$ & 24.3 & 4.3 & 44.1 & 53.2 & 27.9 & 19.8 & 37.6 & 44.1 & 25.3 & 17.0 & 31.1 & 39.2\\
    DistMult$^\dagger$ & 44.4 & 41.2 & 47.0 & 50.4 & 28.1 & 19.9 & 30.1 & 44.6 & - & - & - & -\\
    ComplEx$^\dagger$ & 44.9 & 40.9 & 46.9 & 53.0 & 27.8 & 19.4 & 29.7 & 45.0 & - & - & - & -\\
    RotatE$^\dagger$ & 47.6 & 42.8& 49.2 & 57.1 & 33.8 & 24.1 & 37.5 & 53.3 & 29.0 & 23.4 & 32.2 & 39.0\\
    TuckER$^\dagger$ & 47.0 & 44.3 & 48.2 & 52.6 & 35.8 & 26.6 & 39.4 & 54.4 & - & - & - & -\\
    CompGCN$^*$ & 47.9 & 44.3 & 49.4 & 54.6 & 35.5 & 26.4 & 39.0 & 53.5 & - & - & - & -\\
    BKENE$^*$ & 48.4 & 44.5 & 51.2 & 58.4 & 38.1 & 29.8 & \underline{42.9} & 57.0 & - & - & - & -\\  
    CompoundE$^*$ & 49.1 & 45.0 & 50.8 & 57.6 & 35.7 & 26.4 & 39.3 & 54.5 & - & - & - & -\\
    SymCL$^*$ & 49.1 & 44.8 & 50.4 & 57.6 & 37.1 & 27.6 & 41.1 & 56.6 & - & - & - & -\\ 
    MGTCA$^*$ & 51.1 & 47.5 & 52.5 & 59.3 & \underline{39.3} & 29.1 & 42.8 & \textbf{58.3} & - & - & - & -\\ 
    \hline
    \multicolumn{13}{c}{PLM-based Methods} \\
      \hline
    KG-BERT$^*$ & - & - & - & 52.4 & - & - & - & 42.0 & - & - & - & -\\
    MTL-KGC$^*$ & 33.1 & 20.3 & 38.3 & 59.7 & 26.7 & 17.2 & 29.8 & 45.8 & - & - & - & -\\
    StAR$^*$ & 40.1 & 24.3 & 49.1 & 70.9 & 29.6 & 20.5 & 32.2 & 48.2 & - & - & - & -\\
    KGT5$^*$ & 50.8 & 48.7 & - & 54.4 & 27.6 & 21.0 & - & 41.4 & - & - & - & -\\
    KG-S2S$^*$ & 57.4 & 53.1 & 59.5 & 66.1 & 33.6 & 25.7 & 37.3 & 49.8 & - & - & - & -\\
    SimKGC$^*$  & 67.1 & 58.5 & 73.1 & 81.7 & 33.3 & 24.6 & 36.2 & 51.0 & 35.3 & 30.1 & 37.4 & 44.8  \\
    SimKGC-SymCL$^*$  & 65.7 & 54.6 & 70.9 & 79.1 & 32.4 & 23.5 & 35.4 & 50.4 & - & - & - & -  \\
    GHN$^*$  & \underline{67.8} & \textbf{59.6} & 71.9 & 82.1 & 33.9 & 25.1 & 36.4 & 51.8 & 36.4 & 31.7 & 38.0 & 45.3 \\
    MPIKGC-S$^*$  & 61.5& 52.8 & 66.8 & 76.9 & 33.2 & 24.5 & 36.3 & 50.9 & - & - & - & - \\
    \hline
 Bridge-TransE&  \textbf{69.4} & \underline{59.4} & \textbf{74.7} & \textbf{85.9} & 38.0  & \textbf{31.6} & 41.2 & 57.4 & \underline{45.4} & \underline{40.2} & \underline{47.8} & \textbf{55.6}\\
    \hline
 Bridge-RotatE & 67.3 & 58.3 & \underline{73.3} & \underline{83.2} & \textbf{40.3} & \underline{31.5} & \textbf{43.2} & \underline{58.1} & \textbf{46.2} & \textbf{41.1} & \textbf{48.3} & \underline{55.2} \\
    \hline  
  \end{tabular}
  }

\label{table:Main}
\end{table*}

\subsection{Baseline}
We compare \our with two categories of baselines in Table \ref{table:Main}. 
\textbf{Structure-based methods} aim to learn entity and relation embeddings by modeling relational structure in KGs. 
\textbf{PLM-based methods} aim to enrich knowledge representation by leveraging the semantic knowledge of PLMs but ignore the structural knowledge of KGs, and disregard the disparity between PLMs and KGs, as PLMs are not trained on KGs.
\subsection{\our Setups}\label{append:setup}

We use the bert-base-uncased model as the initialized encoder. In the fine-tuning PLMs with BYOL module, we train \our-TransE on WN18RR, FB15k-237, and Wikidata5M datasets for 2, 2, and 1 epoch(s), respectively. For \our-RotatE, we conduct training on the WN18RR, FB15k-237, and Wikidata5M datasets for 1, 2, and 1 epoch(s), respectively. The initial learning rates  are $4*10^{-4},3*10^{-5},4*10^{-5}$. In the structural triple knowledge learning module, we train \our-transE for 7, 10, and 1 epoch(s) on the respective datasets and \our-RotatE for 8, 10, and 1 epoch(s). 
The corresponding initial learning rates are $1*10^{-4},1*10^{-5},3*10^{-5}$. 
The batch size, additive margin $\gamma$ of contrastive loss, and the temperature $\tau$ are consistent across all datasets, set as 1024, 0.02, and 0.05, respectively. 
\subsection{Overall Evaluation Results and Analysis}
The performances of all models on three datasets are reported in Table \ref{table:Main}. 
Compared with the best baseline results, the improvements obtained by \our-TransE in terms of MRR,  Hits@3, and Hits@10 are 2.4\%, 2.2\%, 4.6\% on WN18RR. Meanwhile, the improvements obtained by \our-RotatE remain competitive with GHN.
On Wikidata5M-Trans dataset, both \our-TransE and \our-RotatE demonstrate substantial improvements. Compared to the best baseline, GHN, \our-TransE achieves increases of 24.7\% in MRR, 26.8\% in Hits@1, 25.8\% in Hits@3, and 22.7\% in Hits@10. Similarly, \our-RotatE achieves increases of 26.9\% in MRR, 29.7\% in Hits@1, 27.1\% in Hits@3, and 21.9\% in Hits@10, respectively.
On FB15k-237, \our-RotatE achieves the best results in MRR and Hits@3, while \our-TransE exhibits comparable performance to the best baseline results in MGTCA. Considering that FB15k-237 is much denser (average degree is $\sim$ 37 per entity) \cite{wang2022simkgc}, MGTCA likely holds an advantage in utilizing abundant neighboring information for learning entity embeddings.


\begin{table*}[t]
\caption{Ablation study on WN18RR, FB15k-237 and Wikidata5M-Trans.}  
\label{table:ablation}  
\centering
\resizebox{\textwidth}{!}{%
  \begin{tabular}{l||cccc|cccc|cccc}
  \hline
     & \multicolumn{4}{c|}{WN18RR}   & \multicolumn{4}{c|}{FB15k-237}  & \multicolumn{4}{c}{Wikidata5M-Trans} \\
      \cline{2-13}
      Model & {MRR} & {Hits@1} & {Hits@3} & {Hits@10} & {MRR} & {Hits@1} & {Hits@3} & {Hits@10} & {MRR} & {Hits@1} & {Hits@3} & {Hits@10}\\
      \hline
     SimKGC  & 67.1 & 58.5 & 73.1 & 81.7 & 33.3 & 24.6 & 36.2 & 51.0 & 35.3 & 30.1 & 37.4 & 44.8\\
     \hline
    \hline
    w/o structural-TransE & 58.2 & 45.2 & 64.4 & 79.3 & 31.0 & 24.2 & 31.9 & 44.7  & 30.1 & 27.7 & 30.0 & 38.1\\
    w/o BYOL-TransE & 67.3 & 59.0 & 72.2 & 80.8 & 37.2 & 30.5 & 40.8 & 56.4  & 40.6 & 33.8 & 40.2 & 50.6\\
    Bridge-TransE &  69.4 & 59.4 & 74.7 & 85.9 & 38.0  & 31.6 & 41.2 & 57.4 & 45.4 & 40.2 & 47.8 & 55.6\\
    \hline
    \hline
     w/o structural-RotatE &  53.9 & 43.2 & 60.1 & 74.1 & 31.8  & 24.1 & 33.8 & 46.3 & 31.4 & 28.2 & 29.8 & 38.4 \\ 
    w/o BYOL-RotatE &  65.4 & 57.2 & 70.8 & 79.6 & 39.6  & 30.8 & 42.7 & 57.3 & 41.1 & 34.0 & 41.5 & 50.8\\
    Bridge-RotatE & 67.3 & 58.3 & 73.3 & 83.2 & 40.3 & 31.5 & 43.2 & 58.1 & 46.2 & 41.1 & 48.3 & 55.2 \\
     \hline
    \hline   
  \end{tabular}
  }
\end{table*}


\begin{table*}[t] 
  \caption{Case study on the tail entity prediction $(h,r,?)$ task using the test set of Wikidata5M-Trans. The \textbf{Bold} font represents the true tail entity. Top 3 shows the first three tail entities that SimKGC and \our predicted, respectively.} 
\label{table:case}
\centering
\resizebox{\textwidth}{!}{%
  \begin{tabular}{l||cc|cc}
  \hline
     & \multicolumn{2}{c|}{SimKGC}   & \multicolumn{2}{c}{\our}\\
      \cline{2-5}
      Triple & {Rank} & {Top 3} & {Rank} & {Top 3} \\
      \hline  
    \emph{(rio pasion, mouth of the watercourse, \textbf{Usumacinta river})} & 119 & \emph{Golfo de Paria, El Golfo de Guayaquil, Yuma River} & 2 & \emph{Tabasco River, \textbf{Usumacinta river}, tzala river} \\
    \hline
    \emph{(lewis gerhardt goldsmith, instance of, \textbf{Human})} & 11 & \emph{plant death, dispute, internet hoax} & 1 & \emph{\textbf{Human}, Lists of people who disappeared,  Strange deaths} \\
    \hline
    \emph{(cross country championships - short race, sport, \textbf{Athletics})} & 4 & \emph{Cross-country running, long distance race, Road run} & 1 & \emph{\textbf{Athletics}, Tower running, Athletics at the Commonwealth}\\
    \hline
  \end{tabular}
  }
\end{table*}

\subsection{Ablation Study}\label{section:ablation}
 To explore the effectiveness of each module, we conduct two variants of \our: (1) removing the structural Triple Knowledge Learning module (referred to as ``w/o structural-TransE'' and ``w/o structural-RotatE''). For inference, we use the fine-tuned \emph{online} BERT and \emph{target} BERT to encode $(h,r)$ and $t$, respectively, and rank the plausibility of each triple based on their cosine similarity (refer to Eq.\eqref{eq:score_transE} and Eq.\eqref{eq:score_rotatE}); (2) remove the Fine-tuning PLMs with BYOL module (referred to as ``w/o BYOL-TransE'' and ``w/o BYOL-RotatE''). 
 The results are summarized in Table \ref{table:ablation}.


\textbf{Effectiveness of Structured Triple Knowledge Learning:} Compared with \our-TransE and \our-RotatE, the results of ``w/o structural-TransE'' and ``w/o structural-RotatE'' reveal that removing the Structured Triple Knowledge Learning module results in notable decreases. This indicates that contrastive loss 
effectively distinguishes similar yet distinct instances. 
The objective of BYOL is to utilize a non-negative strategy to acquire a good initialization that can be applied in downstream tasks. Negative samples continue to play a crucial role in maintaining high performance in these downstream tasks \cite{kim2022bootstrapped,thakoor2021bootstrapped}.
The limitation of relying solely on BYOL arises from the fact that while the non-negative strategy can effectively minimize the gap between representations of distinct views from the same object, it is unable to sufficiently distinguish and disentangle the representations of views originating from similar yet distinct objects. 

\begin{table*}
\centering
  \caption{Error Analysis on the tail entity prediction $(h,r,?)$  on WN18RR. The \textbf{Bold} represents the true tail entity. Top 3 shows the first three tail entities predicted by \our.} 
\label{table:error}
\small
\resizebox{\textwidth}{!}{
  \begin{tabular}{l||cc}
  \hline
      Triple & {Rank} & {Top 3}  \\
      \hline  
    \emph{(position, hypernym, \textbf{location})} & 3 & \emph{region, space, \textbf{location}}  \\
    \hline
    \emph{(take a breather, derivationally related form, \textbf{breathing time})} & 1 & \emph{\textbf{breathing time}, rest, restfulness}  \\
    \hline
    \emph{(Africa, has part, \textbf{republic of cameroon})} & 14 & \emph{Eritrea, sahara, tanganyika} \\
    \hline
  \end{tabular}
 }
\end{table*}
\textbf{Effectiveness of Fine-tuning PLMs with BYOL:} Comparing with \our-TransE and \our-RotatE, the results of ``w/o BYOL-TransE'' and ``w/o BYOL-RotatE'' reveal that removing the fine-tuning BERT with BYOL module results in notable decreases across all metrics in Wikidata5M-Trans, and a minor decline on both WN18RR and FB15k-237. This phenomenon illustrates the necessity for fine-tuning PLMs. While PLMs utilize vast, unlabeled corpora during training to construct a comprehensive language model that embodies textual content, achieving competitive performance in particular tasks often requires an additional fine-tuning step. The results validate our previous speculation that abundant data is crucial for fine-tuning the model since Wikidata5M-Trans is larger than the other two datasets. Therefore, removing fine-tuning BERT with the BYOL module has a more significant negative impact on Wikidata5M-Trans. 
Compared with SimKGC, ``w/o BYOL-TransE'' and ``w/o BYOL-RotatE'' outperforms on FB15k-237 and Wikidata5M-Trans. On WN18RR, ``w/o BYOL-TransE'' outperforms SimKGC in Hits@1 and MRR while being comparable in Hits@3 and Hits@10. This illustrates that our structural scoring function can effectively reconstruct KG's structures in the semantic embedding. 
\subsection{Case Study}

As shown in Table \ref{table:case}, for the first example, the top three tail entities predicted by \our-TransE are rivers in Mexico and geographically close to the true tail entity \emph{\textbf{Usumacinta river}}. However, the top three tail entities SimKGC predicted are rivers in South America. In the second example, the relation \emph{instance of} has ambiguous semantic interpretations. SimKGC cannot capture the semantics of this relation for this triple from the PLMs, resulting in incorrect predictions for the top three tail entities. \our-TransE can understand this relation from the structural perspective, allowing for better predictions. These two toy examples show that when the semantics of the relations are ambiguous, integrating structural knowledge can help to learn a better relation representation.
In the third example, although \our-TransE predicts the true tail entity \emph{\textbf{Athletics}}, the prediction \emph{Cross-country running} made by SimKGC can be regarded as correct. \emph{Cross-country running} and \emph{\textbf{Athletics}} are not mutually exclusive concepts. However, the evaluation metrics consider it an incorrect answer since the triple \emph{(cross country championships - men's short race, sport, Cross-country running)} is not present in KGs. 

\subsection{Error Analysis}

As shown in Table \ref{table:error}, in the first example, \our-TransE ranks the true tail entity \emph{\textbf{location}} as the third. However, the first two tail entities are correct based on human observation. In the second example, \emph{rest} can also be a valid tail due to the fact that \emph{rest} and \emph{\textbf{breathing time}} are lexically similar concepts. In the third example, \our-TransE ranks the true tail entity \emph{\textbf{republic of cameroon}} as 14th, attributed to the nature of the relation \emph{has part}, which is a many-to-many relation. The first three tail entities predicted by \our-TransE are correct because they are all located in Africa.
Drawing from these observations, some predicted triples might be correct based on human evaluation. However, these triples might not be present in KGs. This false negative issue results in diminished performance. 
\subsection{Efficiency of \our}\label{subsection:efficiency}

We run SimKGC \footnote{https://github.com/intfloat/SimKGC} on WN18RR and conduct an efficiency comparison with \our-TransE. 
Table \ref{table:efficiency} illustrates the model efficiency of \our-TransE and SimKGC on WN18RR with a batch size of 1024. In \our-TransE, the Fine-tuning PLMs with BYOL step converges in 2 epochs, and the Structured Triple Knowledge Learning step achieves convergence in 7 epochs (9 epochs in total). The total training time is 3550 seconds. SimKGC converges in 8 epochs, and the total training time is 3331 seconds. Consequently, the overall computational cost of \our is comparable with SimKGC.
\begin{table}
\centering
\caption{Model efficiency of \our-TransE and SimKGC on WN18RR.} 
  
\label{table:efficiency}
\resizebox{0.48\textwidth}{!}{
  \begin{tabular}{c|c|c}
  \hline
      Model & \# Total Training Epoch & \# Total Training Time  \\
      \hline  
   SimKGC & 8 & 3331s \\
    \hline
    \our-TransE & 9 & 3550s \\
    \hline
  \end{tabular}
}
 
\end{table}
  

\section{Conclusion}\label{sec:conclusion}
In this paper, we introduce \our, which integrates PLMs with structure-based models. 
Since no previous study investigates structural principles using
PLM-based representation, we jointly encode structural and semantic information of KGs to enhance knowledge representation. Further, existing work overlooks the gap between KGs and PLMs due to the absence of KG training in PLMs. To address this issue, we utilize BYOL to fine-tune PLMs. 
Experimental results demonstrate \our outperforms most baselines. 


\noindent \textbf{Acknowledgement.} The work is supported in part by NSF-CAREER 2237831. 

\bibliographystyle{splncs04}
\bibliography{reference}
%




\end{document}